\documentclass[journal,twoside,web]{ieeecolor}
\usepackage{generic}
\usepackage{cite}
\usepackage{amsmath,amssymb,amsfonts}
\usepackage{algorithmic}
\usepackage{graphicx}
\usepackage{textcomp}
\usepackage{caption}
\usepackage{booktabs} 
\usepackage{multirow} 
\usepackage{tabularx}
\usepackage{arydshln} 
\usepackage{hyperref}
\newcolumntype{Y}{>{\centering\arraybackslash}X} 
\newcolumntype{A}[1]{>{\hsize=#1\hsize\centering\arraybackslash}X} 

\def\BibTeX{{\rm B\kern-.05em{\sc i\kern-.025em b}\kern-.08em
    T\kern-.1667em\lower.7ex\hbox{E}\kern-.125emX}}
\begin{document}
\title{Lightweight Method for Interactive 3D Medical Image Segmentation with Multi-Round Result Fusion}
\author{Bingzhi Shen, Lufan Chang$^{*\dag}$, Siqi Chen, Shuxiang Guo$^{*}$ \IEEEmembership{Fellow, IEEE}, Hao Liu
\thanks{Manuscript received December 2024; This work was supported by Yizhun Medical AI Co., Ltd,.(This research was conducted while Bingzhi Shen and Siqi Chen were doing internships at Yizhun Medical AI.)}
\thanks{Bingzhi Shen is with the Aerospace Center Hospital, School of Life Science and the Key Laboratory of Convergence Medical Engineering System and Healthcare Technology, Ministry of Industry and Information Technology, Beijing Institute of Technology, Beijing 100081, China (e-mail: shenbz@bit.edu.cn).}
\thanks{Lufang Chang and Hao Liu are with the Yizhun Medical AI Co., Ltd, Beijing, 100081, China (email: lufan.chang@yizhun-ai.com, hao.liu@yizhun-ai.com).}
\thanks{Siqi Chen is with the Shenzhen Key Laboratory of Ubiquitous Data Enabling, Tsinghua Shenzhen International Graduate School, Tsinghua University, Shenzhen, China (email: csq23@mails.tsinghua.edu.cn).}
\thanks{Shuxiang Guo is with the Department of Electronic and Electrical Engineering, Southern University of Science and Technology, Shenzhen, Guangdong 518055, China, and also with the Aerospace Center Hospital, School of Life Science and the Key Laboratory of Convergence Medical Engineering System and Healthcare Technology, Ministry of Industry and Information Technology, Beijing Institute of Technology, Beijing 100081, China (e-mail:guo.shuxiang@sustech.edu.cn).}
\thanks{\dag: Project Lead: Lufan Chang.}
\thanks{*: Corresponding authors: Lufan Chang, Shuxiang Guo.}
}
\maketitle

\begin{abstract}
In medical imaging, precise annotation of lesions or organs is often required. However, 3D volumetric images typically consist of hundreds or thousands of slices, making the annotation process extremely time-consuming and laborious. 
Recently, the Segment Anything Model (SAM) has drawn widespread attention due to its remarkable zero-shot generalization capabilities in interactive segmentation. While researchers have explored adapting SAM for medical applications, such as using SAM adapters or constructing 3D SAM models, a key question remains: \textsc{can traditional CNN networks achieve the same strong zero-shot generalization in this task?}

In this paper, we propose the \emph{L}ightweight \emph{I}nteractive Network for 3D \emph{M}edical Image Segmentation (LIM-Net), a novel approach demonstrating the potential of compact CNN-based models. 
Built upon a 2D CNN backbone, LIM-Net initiates segmentation by generating a 2D prompt mask from user hints. This mask is then propagated through the 3D sequence via the Memory Module. To refine and stabilize results during interaction, the Multi-Round Result Fusion (MRF) Module selects and merges optimal masks from multiple rounds.

Our extensive experiments across multiple datasets and modalities demonstrate LIM-Net's competitive performance. It exhibits stronger generalization to unseen data compared to SAM-based models, with competitive accuracy while requiring fewer interactions. Notably, LIM-Net's lightweight design offers significant advantages in deployment and inference efficiency, with low GPU memory consumption suitable for resource-constrained environments. These promising results demonstrate LIM-Net can serve as a strong baseline, complementing and contrasting with popular SAM models to further boost effective interactive medical image segmentation. The code will be released at \url{https://github.com/goodtime-123/LIM-Net}.

\end{abstract}

\begin{figure*}[!t]
  \centering
  \includegraphics[width=20cm, height=12cm, keepaspectratio]{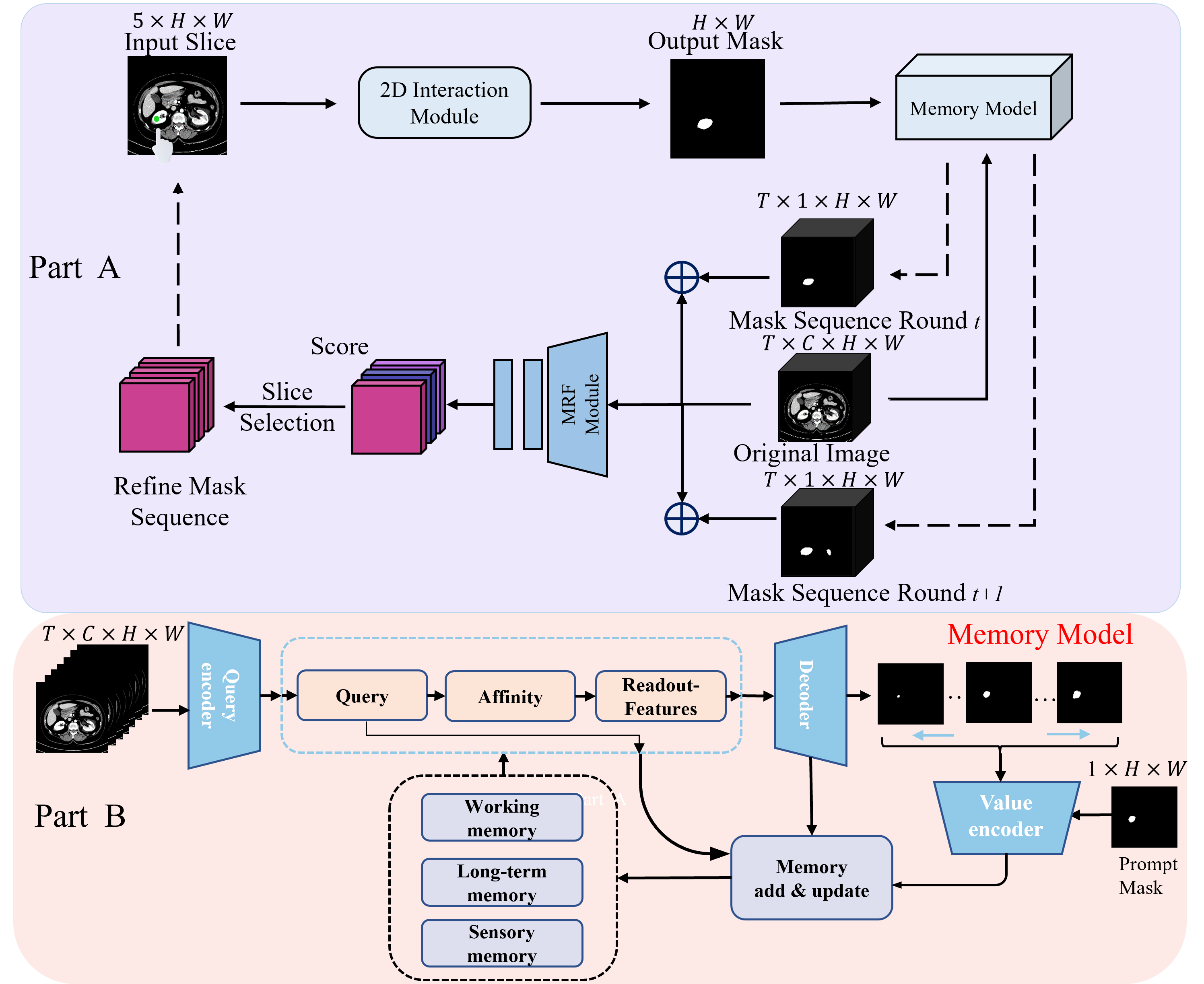}
  \caption{
The proposed LIM-Net consists of two parts: (Part A) It first generates initial masks from user clicks using a 2D Interaction Module. Then, a memory model extends the prompt mask to the entire sequence. Finally, MRF module select mask across multiple interaction rounds for consistent segmentation quality. (Part B) a Memory Model that incorporates different types of memory to retrieve relevant information and generate masks by single prompt mask.
}
  \label{fig1}
\end{figure*}

\section{Introduction}
\label{sec:introduction}

\IEEEPARstart{M}{edical} image segmentation plays an important role in various clinical applications. While supervised learning have made significant advancements in this field, it relies heavily on large-scale annotated datasets. Volumetric medical scans, such as computed tomography (CT) and magnetic resonance imaging (MRI), typically comprise numerous cross-sectional slices, making the manual annotation process exceptionally labor-intensive and time-consuming, even for domain experts.
Fully automated segmentation approaches \cite{isensee2021nnunet,hatamizadeh2021swinunetr,shaker2022unetr++,zhou2023nnformer} have been researched but can struggle with complex scenarios like ambiguous lesion boundaries or intricate anatomical structures. To alleviate this burden, interactive segmentation techniques that can provide accurate results while minimizing user effort have garnered significant interest. To address this challenge, interactive image segmentation that can yield accurate results while minimizing user input have attracted considerable research interest.

The Segment Anything Model (SAM) \cite{kirillov2023sam} has gained attention due to its remarkable zero-shot generalization capabilities in interactive segmentation across visual domains. SAM utilizes a Vision Transformer (ViT) encoder \cite{dosovitskiy2020vit} to extract semantic feature representations from the input images. A prompt encoder module encodes the user's interaction hints (e.g., points or bounding boxes) into compact embedding vectors. The mask decoder integrates these elements through self-attention and cross-attention mechanisms, ultimately generating the corresponding segmentation mask predictions. SAM has demonstrated superior zero-shot ability, where the model can produce accurate segmentation results solely relying on rough object indications. While researchers have explored adapting SAM for 3D medical applications, such as using SAM adapters \cite{wu2023medicalsamadapter,gong20233dsamadapter} or constructing 3D SAM models \cite{wang2023sammed3d}, these approaches do not yet provide competitive results ready for real-world clinical scenarios. One possible underlying reason could be that Transformers often lack some of the inductive biases inherent to CNNs, such as translation equivariance and locality, and therefore do not generalize well when trained on insufficient amounts of data \cite{dosovitskiy2020vit}. Furthermore, 3D SAM model incorporate more parameters, complicating training and increasing inference time. This heightened computational complexity impedes deployment in resource-limited clinical environments. Studies have shown that CNNs may struggle with generalization to out-of-distribution samples \cite{bai2021transformersrobust,wang2018imagefinetune}. However, medical images possess unique characteristics that differ from natural images. In radiological imaging and ultrasound, the regions of interest typically exhibit distinct intensity variations. Even in surgical video data, the objects to be segmented are generally well-defined. We hypothesize that, if properly designed and trained, CNN-based approaches have the potential to enable real-time interactive medical image segmentation while maintaining accuracy and exhibiting robust generalization to unseen data.
Owing to real-time performance and memory efficiency requirements, previous CNN-based interactive segmentation models have predominantly relied on 2D slice-based propagation instead of heavy 3D convolutions. Video Object Segmentation (VOS) have emerged as a promising solution, as they enable rapid and efficient segmentation across video frames by leveraging temporal consistency.

Since volumetric images can be represented as sequences of 2D slices, researchers have developed interactive 3D segmentation algorithms based on VOS \cite{zhou2021qualityawaremem,shi2022hybridpropagat,liu2023CycleConsistency} to effectively track and segment objects. These approaches typically decouple the task into two components: user hint integration for initial mask generation, and cross-slice propagation guided by the volumetric data.
Notably, the Quality-Aware Memory Network \cite{zhou2021qualityawaremem} is built upon the Space-Time Correspondence Network (STCN)  \cite{cheng2021rethinkingvideoseg}. It incorporates a quality assessment module to assist users in selecting informative slices. Shi et al. \cite{shi2022hybridpropagat} propose a hybrid network combining 2D and 3D CNNs to extract multi-scale features from volumetric data. They typically require retraining on the target dataset \cite{zhou2021qualityawaremem,wang2018deepigeos,shi2022hybridpropagat} or at least fine-tuning on the target image \cite{wang2018imagefinetune}. This deficiency stems from the data-hungry nature of CNN models and their inability to effectively leverage prior knowledge when confronted with new data distributions. While slice-based propagation enables 3D volume processing, its efficacy diminishes for extended sequences of hundreds or thousands of slices. This decline is primarily due to the increasing difficulty in modeling long-range dependencies between frames. Model accuracy typically decreases as the distance from the prompt frames increases towards both ends of the sequence, reflecting the challenge of capturing frame-to-frame relationships over longer ranges.

To address these issues, we propose LIM-Net, a novel CNN-based model for interactive 3D medical image segmentation. LIM-Net is trained on a large-scale dataset encompassing diverse imaging modalities and anatomical structures, enhancing its generalization capabilities and enabling robust feature acquisition for novel scenarios.
Inspired by the XMem \cite{cheng2022xmem}, LIM-Net incorporates a hierarchical memory module with three levels of memory stores to effectively capture both short and long-range dependencies across extensive volumetric sequences, strengthening its long-term memory and modeling capacity.
We further introduce a MRF Module to selectively merge and refine predictions from multiple interaction rounds, improving segmentation consistency across 3D volumes, particularly for slices distant from user interaction slice. Through these approaches, LIM-Net aims to provide accurate and robust interactive 3D medical image segmentation while maintaining efficiency for resource-constrained clinical environments. The following sections detail the architecture of LIM-Net and comprehensively evaluate its performance across multiple datasets and evaluation metrics.

\subsection{Contributions}
In summary, the primary contributions of this study are outlined as follows:

1. We propose LIM-Net, an advanced long-term 3D interactive medical image segmentation framework featuring real-time performance and accurate segmentation of lengthy sequences.


2. An innovative MRF module is designed to stabilize and improve segmentation quality for distant slices by merging multi-round predictions.


3. A defective mask generation strategy using random transformations to enhance the robustness and interaction responsiveness of the segmentation modules, addressing the insensitivity to user prompt points in previous methods.

4. LIM-Net is trained on a large-scale diverse dataset, demonstrating strong generalization capability on unseen data through extensive experiments.


\section{METHODOLOGY}
As shown in Figure~\ref{fig1}, LIM-Net comprises three key components. The details of each component will be elaborated in the following sections.



\subsection{Problem Setting}
We focus on the task of accurately segmenting target objects (e.g., organs or lesions) in 3D volumetric images. Let \( s_i \) denote the \( i \)-th 2D slice of a volumetric image \( V \), where \( V \) consists of \( N \) slices, and \( 1 \leq i \leq N \). The original image \( V \) can be represented as \( V = \{ s_1, s_2, \ldots, s_N \} \).
During segmentation, the user selects an arbitrary slice $s_k$ ($1 \leq k \leq N$) and interactively annotates the target object using click-based annotation. Through a 2D interaction module, we obtain the segmentation result $m_k$.
Our goal is to propagate the segmentation result of the single slice $m_k$ to infer the segmentation masks $M = {m_1, m_2, \ldots, m_N}$ for the entire volumetric image $V$, where $m_i$ is the mask for slice $s_i$.

\subsection{Prompt Mask Generation}

Following the setting of RITM \cite{sofiiuk2022ritm}, the user-annotated points are encoded into click maps through a click encoder. These click maps are then added to the image feature maps, which are subsequently fed into the HRNet-32 backbone \cite{wang2020hrnet} to produce the initial 2D prompt mask. To tailor the 2D interaction module for medical image segmentation, we trained it on a collection of diverse 2D medical datasets. This strategy effectively enhances the network's capability to generate accurate masks from user hints on individual 2D slices, before extending it to model full 3D sequences.


\setlength{\heavyrulewidth}{1.2pt} 
\setlength{\lightrulewidth}{0.6pt} 
 \setlength{\tabcolsep}{10mm}
\begin{table*}[ht]
\centering
\caption{Evaluation Results(DSC, \%) of Various Methods on the KiTS19 and MSD Datasets}
\label{tb1}
\begin{tabular}{@{}lcccc@{}}
\toprule
Methods            & KIDNEY (Organ) & KIDNEY (Tumor) & Lung & Colon  \\ \midrule
nnIU-Net (6 points)\cite{zhou2021qualityawaremem}           & 94.5           & 86.3           & 73.9        & 68.1         \\
DeepIGeoS (6 points)\cite{zhou2021qualityawaremem}          & 95.7           & 87.6           & 76.6        & 72.3         \\
Mem3d (6 points)\cite{zhou2021qualityawaremem}              & 96.9           & 88.2           & 80.9        & 79.7         \\
Ours w/o MRF       &  96.8              &84.8                &83.0             &77.1              \\

\midrule
Ours-MRF(6 points)               & \textbf{97.3}          & \textbf{89.2}          & \textbf{83.3}       & \textbf{81.4}        \\ \bottomrule
\end{tabular}
\end{table*}


%
\subsection{Multi-Round Fusion}
\begin{figure}[!t]
\centering
\includegraphics[width=0.48\textwidth]{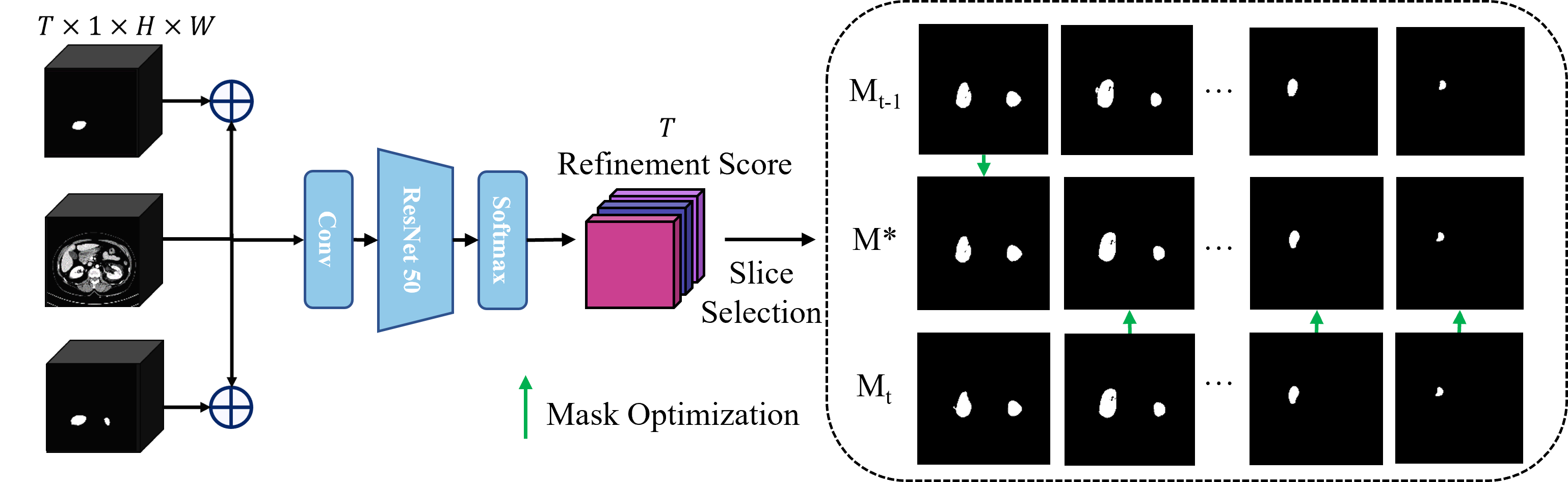}
\caption{The overall framework of the MRF module.}
\label{mrf2}
\vspace{-0.5cm} 
\end{figure}

To mitigate inconsistent slice-level segmentation updates from single-round user interaction, especially for distant slices, we propose a Multi-Round Result Fusion (MRF) module, as shown in the figure \ref{mrf2} . Instead of solely relying on the latest round's output, MRF selectively fuses high-quality masks from the current and previous interaction rounds for robust 3D segmentation results.
\subsubsection{Problem Formulation}
Given two complete sets of 2D segmentation masks from consecutive interaction rounds, $M^{t-1}$ and $M^t$, our goal is to generate an optimal fusion $M^*$ that maximizes segmentation quality across all slices. The fusion problem can be formulated as:
\begin{equation}
M^*=\arg\max_M\sum_{i=1}^N\left(P_i\cdot\delta(m_i^{t-1})+(1-P_i)\cdot\delta(m_i^t)\right)
\label{eq:optimization_max}
\end{equation}
where $M = \{ M_i \}_{i=1}^N$ denotes the set of fused segmentation results to be optimized. $P_i$ is the probability that the previous mask $m_i^{t-1}$ is of higher quality than the current mask $m_i^t$, as determined by the quality assessment network.  
$\delta(m_i^{t-1})$ and $\delta(m_i^t)$ are indicator functions that equal 1 if $M_i^*$ equals $m_i^{t-1}$ and $m_i^t$ respectively, and 0 otherwise.
\subsubsection{Quality Assessment Network}
The core of our MRF module is a ResNet-50-based quality assessment network that evaluates the relative quality of segmentation masks from different rounds. Specifically, starting from the second round of interaction, we have access to two complete 2D mask sets: $M^{t-1}$ from the previous round and $M^t$ from the current round, where $M^t$ is newly generated by the re-segmentation process.
The user clicks on a slice $s_i$ within $V$, which produces
segmentation mask $M^t$. Although $M^t$ generally outperforms $M^{t-1}$, it does not guarantee improved segmentation quality for every individual slice.
For each slice $s_i$, we compute a quality score:
\begin{equation}
P_i = f\left(s_i, m_i^{t-1}, m_i^t \right)
\label{eq1}
\end{equation}
where $f(\cdot)$ represents our quality assessment network. The network processes the concatenated input of the original image slice and both mask versions to produce a single probability value $P_i \in [0,1]$. This probability indicates whether the previous round's mask ($m_i^{t-1}$) is of better quality than the current round's mask ($m_i^t$).

In practice, to enhance computational efficiency, instead of processing individual slices sequentially, we implement batch processing for the quality assessment. The network processes multiple slices simultaneously, which can be expressed as:
\begin{equation}
\mathbf{P} = f(V, M^{t-1}, M^t\}; \theta)
\end{equation}
where $V = {s_1, s_2, ..., s_b}$ represents a batch of $b$ image slices, $M^{t-1} = {m_1^{t-1}, m_2^{t-1}, ..., m_b^{t-1}}$ and $M^t = {m_1^t, m_2^t, ..., m_b^t}$ are their corresponding masks from previous and current rounds, respectively. The network outputs a batch of probability scores $\mathbf{P} = {P_1, P_2, ..., P_b}$ simultaneously. The network parameters are denoted by $\theta$. This batch processing approach significantly reduces computational overhead and accelerates the overall inference process compared to slice-by-slice evaluation. The batch size $b$ can be adjusted based on available GPU memory and computational resources to optimize the trade-off between processing speed and memory consumption.
\subsubsection{Fusion Strategy}
The fusion process operates on a slice-by-slice basis using a straightforward decision rule:
\begin{equation}
M_i^* = \begin{cases}
m_i^{t-1} & \text{if } P_i > \tau \\
m_i^t & \text{otherwise}
\end{cases}
\end{equation}
where $M_i^*$ represents the final selected mask for slice $i$. This binary decision threshold of $\tau$ determines whether to retain the previous round's segmentation or accept the new one. This selective fusion strategy ensures that high-quality segmentations from previous rounds are preserved while incorporating improvements from the current round.
The MRF framework provides several key advantages: (1) it maintains segmentation consistency across slices by leveraging information from multiple rounds, (2) it reduces the required number of user interactions by effectively utilizing historical segmentation results, and (3) it guarantees monotonic improvement in overall segmentation quality through intelligent fusion. 
\subsection{Memory-Augmented 3D Sequence Propagation}
This paper presents a fast long-term 3D interactive image segmentation method based on XMem\cite{cheng2022xmem}, aiming to tackle the challenges of low accuracy, high GPU memory consumption, and slow prediction speed in medical image segmentation. The framework of Memory-Augmented Sequence Propagation module is shown in the part B of Figure ~\ref{fig1}. 
In previous designs, users were required to initiate modifications starting from the first slice in the sequence of a Volumetric image, which proved to be inflexible and inefficient. 
To address this issue, we have introduced a bidirectional propagation mechanism. This mechanism treats the prompt mask as the starting point of two separate sequences - one moving forward and the other moving backward. By invoking the model twice, once for each direction, we can efficiently obtain the predicted masks for all frames in the volume. This allows users to initiate the segmentation from any arbitrary slice \( s \) and obtain the masks for the entire sequence, greatly enhancing the flexibility and efficiency of the segmentation process. Let $m_p$ denote the prompt mask. Let $s_1, s_2, \ldots, s_{p-1}$ and $m_1, m_2, \ldots, m_{p-1}$ represent the slices and corresponding masks before the prompt mask, respectively. Similarly, let $s_{p+1}, s_{p+2}, \ldots, s_N$ and $m_{p+1}, m_{p+2}, \ldots, m_N$ represent the slices and corresponding masks after the prompt mask, respectively. Our bidirectional propagation mechanism can be formulated as follows:
\begin{equation}
m_{p+i} = XMem(s_{p+i}, m_p),  i = 1, \ldots, N-p
    \label{eq12}
    \end{equation}
    \begin{equation}
m_{j}  = XMem(s_j, m_p), j = 1, \ldots, p-1
    \label{eq13}
    \end{equation}

This bidirectional approach ensures that each frame, whether before or after the prompt mask \( m_p \), is effectively integrated using the \( XMem \) mechanism, thereby generating a complete set of masks for all slices \( \{ s_1, s_2, \ldots, s_N \} \) in the sequence.

The key components of the method include Sensory Memory, Working Memory, and Long-term Memory.

\textbf{Sensory Memory} focuses on short-term memory to complement the deficiencies in Working and Long-term Memory. 

\textbf{Working Memory} stores the recent high-resolution features to increase matching accuracy. It is composed of $ K^W \in \mathbb{R}^{C^K \times T \times H \times W} $ and $ V^W \in \mathbb{R}^{C^V \times T \times H \times W} $, where T is the number of slices in the Working Memory. Here, $f_{QE}$ and $f_{VE}$ denotes the Query Encoder and Value Encoder, respectively.
The values of $K^W$ and $V^W$ are computed based on the formulas \ref{eq3} and \ref{eq4}.
\begin{equation}
K^W = f_{QE}(Image)
\label{eq3}
\end{equation}
\begin{equation}
V^W = f_{VE}(Image, PredMask)
\label{eq4}
\end{equation}
The computed $K^W$and $V^W$ are added to the working memory and utilized in subsequent memory slice readings.

To control memory usage, the number of slice features in the working memory is constrained within the range 
\begin{equation}
T:T_{min}\le T<T_{max}
\label{eq5}
\end{equation}
where $T_{min}$ and $T_{max}$ represent the minimum and maximum capacity of the working memory respectively. 

\textbf{Long-Term Memory} consolidates features from working memory using a prototype selection strategy, effectively processing long volumetric images by selectively storing key feature.
Memory consolidation\cite{cheng2022xmem} is required when the working memory reaches a predefined size. The slices with user interaction information and the recent $T_{min}-1$ memory slices are kept in the working memory as a high-resolution buffer, while the remaining $(T_{max}-T_{min})$ slices are converted into candidate slices for the long-term memory. These candidate keys and values are denoted as $K^C\subset K^W$ and $V^C\subset V^W$ respectively.
\setlength{\abovedisplayskip}{5pt}
A key selection strategy is utilized to choose high information density $K^P$ and generate associated values $V^P$. For brevity, the details of this strategy are omitted here but you can find more details in XMem\cite{cheng2022xmem}.
The computation process for $V^P$ is as follows. First, the similarity matrix A is computed:
\setlength{\abovedisplayskip}{5pt}
\begin{equation}
S\left(K^C,K^P\right)\ =-s_i\sum_{c}^{C^k}{e_{cj}{({K^C}_{ci}-{K^P}_{cj})}^2}    
\label{eq6}
\end{equation}
Next, the cumulative total affinity matrix W is calculated:
\begin{equation}
{W\left(k,q\right)}_i = \frac{\exp\left({A(k,q)}_i\right)}{\sum_{j=1}^{K}{\exp\left({A(k,q)}_j\right)}}
\label{eq7}
\end{equation}
Using the results from the first two equations, \( V^P \) is computed:
\begin{equation}
V^P = v^cW(k^c, k^P)
\label{eq8}
\end{equation}
Finally, the selected \( K^P \) and computed \( V^P \) are added to the long-term memory \( K^{LongTerm} \) and \( V^{LongTerm} \).


 \begin{figure*}[!t]
  \centering
  \includegraphics[width=12cm, height=8cm, keepaspectratio]{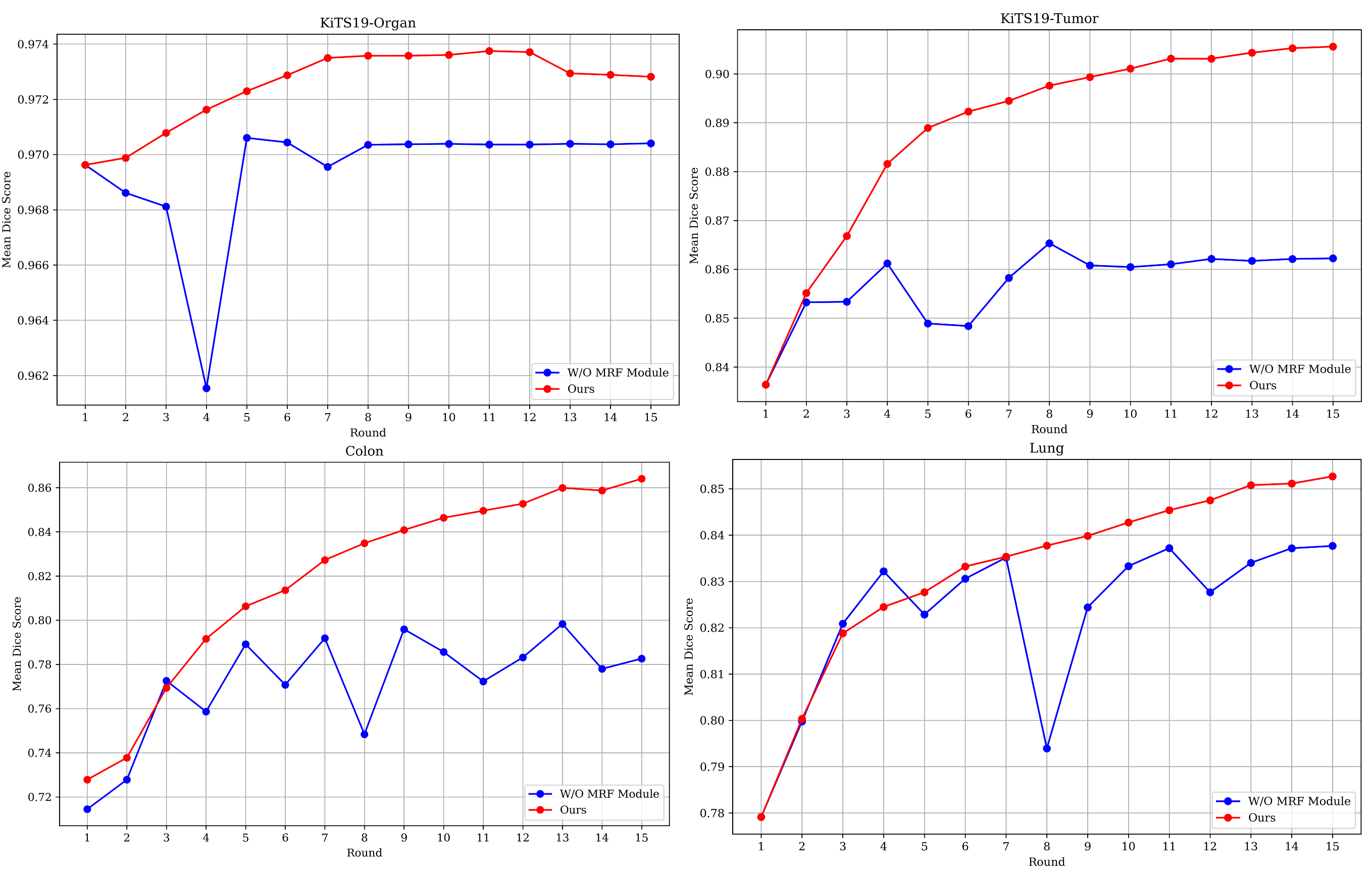}

\caption{Average Dice changes across different interaction rounds on the MSD-lung, MSD-colon, KiTS-Organ, and KiTS-Tumor datasets}
\label{fig3}
\end{figure*}

\section{EXPERIMENTS AND RESULTS}

\subsection{Datasets}
The proposed framework was evaluated using diverse datasets spanning multiple imaging modalities and anatomical regions for training and testing purposes. Table \ref{dataset_detail} presents the dataset details.
 
For training the model, the following datasets were utilized:
\textbf{Ishape}(irregular Shape)\cite{yang2021ishape}: An Irregular Shape Instance Segmentation dataset comprising highly irregular object shapes, with subsets of real and synthetic images. For this study, a single instance was randomly selected and cropped from each image. 
\textbf{In-house CT and ultrasound (US) Dataset}: Our proprietary database contains 845 high-quality US images and 3,323 two-dimensional CT slices transformed from volumetric CT scans. Both datasets were careful annotated by our experienced in-house medical imaging specialists. To ensure annotation accuracy and consistency. \textbf{AMOS} (Abdominal Multi-Organ Segmentation)\cite{ji2022amos}: Provided by the MICCAI 2022 Multi-Modal Abdominal Multi-Organ Segmentation Challenge, this dataset consists of 500 CT and 100 MRI scans with voxel-level annotations for 15 abdominal organs across diverse vendors, modalities, and pathologies. In this study, volumetric data was converted to 2D slices for utilization.
\textbf{DAVIS-585}\cite{chen2022focalclick}:  It comprises 585 test samples derived from 30 validation videos, with 10 frames sampled per video.
\textbf{COCO-LVIS}: Constructed from COCO\cite{lin2014microsoft} and LVIS\cite{gupta2019lvis} datasets, this large vocabulary instance segmentation dataset boasts high annotation quality, comprising 104,000 images and 1.6 million instance masks.
\textbf{MSD-Colon}\cite{simpson2019large}: Consisting of 126 colon cancer CT cases, divided into a training set of 100 cases and a test set of 26 cases.
\textbf{MSD-Lung}\cite{simpson2019large}: A lung cancer subset with a training set of 64 cases and a test set of 32 cases.
\textbf{KiTS19}\cite{heller2019kits19}: Containing 300 abdominal CT scans, with 210 released for benchmarking, further split into a training set of 168 scans and a test set of 42 scans.
 
To assess the model's generalization capabilities, the following datasets were utilized: MSD-Colon, MSD-Lung, KiTS19; detailed information is mentioned before.
 The EndoVis 18 (Endoscopic Vision challenge at MICCAI 2018)\cite{allan2018RoboticScene2020} dataset, consisting of endoscopic images, is used exclusively for testing. Additionally, the MSD-Pancreas\cite{simpson2019large} and BTCV\cite{landman2015miccai} datasets are used only for testing.
To assess model generalization, both seen and unseen datasets were utilized. This combination evaluates the model's ability to generalize to new data, prevents overfitting, and simulates real-world clinical deployment. The diverse imaging modalities (CT, endoscope images) and anatomical regions (colon, lung, kidney, pancreas) provide comprehensive assessment across various scenarios. Seen datasets enable benchmarking against previous methods, while unseen datasets investigate domain adaptation capabilities, potentially guiding transfer learning and training strategies. While seen datasets allow for direct performance comparison with existing methods, testing on unseen datasets helps evaluate the model's true generalization capabilities across different domains.
\setlength{\heavyrulewidth}{1.2pt} 
\setlength{\lightrulewidth}{0.6pt} 
 \setlength{\tabcolsep}{10mm}
\begin{table*}[ht]
\centering
\caption{Effectiveness Analysis of the MRF Module}
\label{tb5}
\begin{tabular}{@{}lcccc@{}}
\toprule
Methods            & KIDNEY (Organ) & KIDNEY (Tumor) & Lung & Colon  \\ \midrule
Ours w/o MRF       &  96.8              &84.8                &83.0             &77.1              \\
Ours-MRF               & \textbf{97.3}          & \textbf{89.2}          & \textbf{83.3}       & \textbf{81.4}    \\  \bottomrule

\end{tabular}
\end{table*}
\subsection{Implementation details}
This study utilized the same parameters and settings as RITM \cite{sofiiuk2022ritm} and XMem\cite{cheng2022xmem}. The MRF module is based on a pretrained ResNet-18 model, which was trained on a defect dataset. The defect dataset was constructed using a baseline model without MRF on four datasets: KiTS19-Organ, KiTS19-Tumor, MSD-lung, and MSD-colon, which comprised both defect masks and ground truth masks.

\textbf{A defective mask generation strategy:} To simulate real-world scenarios, we corrupted the ground truth masks to generate defective masks for training. This approach enhanced the robustness of the 2D interaction module. Furthermore, it improved the responsiveness of the 2D interaction module to user prompts. The corruption process involved applying various random transformations to the original masks, with each type of transformation having a predefined probability of [0.1, 0.1, 0.3, 0.2, 0.1, 0.2].
One of the main corruptions was adding random shapes, such as rectangles, triangles, and polygons, to the masks. The size and position of these shapes were determined based on the contours and area of the original masks, ensuring that the added shapes were of diverse sizes and placed randomly within or near the mask region.
Another type of corruption involved applying morphological operations, like dilation and erosion, to the masks. The number of iterations for these operations was randomly selected, with dilation iterations ranging from 10 to 30 and erosion iterations fixed at 10. 
Additionally, we introduced perturbations to the original mask boundaries by randomly displacing boundary pixels. The maximum displacement distance was randomly chosen between 10 and 30 pixels, creating irregular and jagged edges resembling real-world segmentation challenges.
Other corruptions included randomly smoothing the masks, removing existing shapes from the masks, and merging random shapes near the mask boundaries. The smoothing operation helped to simulate blurred or imprecise segmentation results, while the removal and merging of shapes mimicked the effects of under- and over-segmentation, respectively.
This approach aimed to improve the 2D Interaction module's ability to refine and correct imperfect segmentation results in practical applications. The stochastic nature of the corruption process ensured exposure to diverse, realistic segmentation challenges during training.

For the 2D Interactive module, a multi-stage training strategy was conducted, initiating with weights derived from the RITM-HRNet-W32 model weight. We employed a multi-stage training strategy to progressively refine the models' capabilities and ensure a smooth transition from general image understanding to specialized medical imaging tasks. The 2D models for Tables \ref{exp_btcv}, \ref{tbpancreas}, and \ref{tbendo} were initially trained on a combination of the Ishape \cite{yang2021ishape}, AMOS \cite{ji2022amos}, DAVIS-585 \cite{chen2022focalclick}, COCO-LVIS \cite{lin2014microsoft, gupta2019lvis} , and an In-house CT and US in Stage 1. During Stage 2, the models were trained using a mix of AMOS \cite{ji2022amos}, In-house CT and US, MSD-Colon, MSD-Lung \cite{simpson2019large}, and KiTS19 \cite{heller2019kits19} datasets, thereby ensuring a comprehensive encounter with a wide range of medical imaging data. By utilizing a diverse set of datasets for training, the models were able to learn a wide range of features, enhancing their ability to generalize and adapt to various medical imaging tasks.

To maintain consistency with the experimental setup of the comparative methods in Table \ref{tb1}, which are divided into training and testing sets, the 2D model for Table \ref{tb1} followed a training process. Following Stage 2, Stage 3 further trained the model on these specific datasets, optimizing its performance for the specific dataset. To evaluate the models' generalization capabilities, we tested on unseen datasets, including EndoVis 18 \cite{allan2018RoboticScene2020} , MSD-Pancreas\cite{simpson2019large}, and BTCV \cite{landman2015miccai}. This combination of unseen datasets allowed us to assess the models' robustness, real-world applicability.

The 3D model for Table \ref{tbendo} used XMem-s012 because the domain gap between EndoVis 18 and medical image datasets was substantial. XMem-s012, being a more general-purpose dataset, helped bridge this gap and improve the model's performance on the EndoVis 18 dataset. The 3D model for Table \ref{tb1} was trained on the MSD-Colon, MSD-Lung, and KiTS19 datasets.
we aimed to develop models that can handle a wide range of tasks while excelling in their specific domains, striking a balance between generalizability and specialization.

All models were developed within the Pytorch framework and trained on 8 NVIDIA GeForce RTX 3090 GPUs.

\setlength{\heavyrulewidth}{1.2pt} 
\setlength{\lightrulewidth}{0.6pt} 
 \setlength{\tabcolsep}{10mm}
\begin{table}[ht]
\centering
\caption{Datasets for Constructing the LIM-Net}
\label{dataset_detail}
\begin{tabular}{@{}l@{\hspace{10pt}}c@{\hspace{10pt}}c@{\hspace{10pt}}c@{}}
\toprule
Dataset     &Type       &Training Set  &Test Set  \\ \midrule
Ishape\cite{yang2021ishape}     &Synthesis image        &10,185 &2,685             \\
AMOS\cite{ji2022amos}    &CT             &6,263      &253     \\
MSD-Colon\cite{simpson2019large}     &CT               &1,034 &251         \\
MSD-Lung\cite{simpson2019large}  &CT              &1,140  &517       \\
KiTS19\cite{heller2019kits19}      &CT              & 14,072 & 1,753             \\
COCO-LVIS\cite{lin2014microsoft}\cite{gupta2019lvis}      &Natural image   & 99,354 &4,751           \\
DAVIS-585 \cite{chen2022focalclick}&Natural image   &585  &585  \\  
In-house US dataset   &US      & 683     &171             \\
In-house CT dataset   &CT      & 3,323     &-             \\
BTCV\cite{landman2015miccai} &CT              & - & 2,210 \\
EndoVis18\cite{allan2018RoboticScene2020}   &Endoscope  & - &4,621\\
\bottomrule


\end{tabular}
\end{table}

\setlength{\heavyrulewidth}{1.2pt} 
\setlength{\lightrulewidth}{0.6pt} 
 \setlength{\tabcolsep}{10mm}
\begin{table*}[ht]
\centering
\caption{Evaluation Results(DSC, \%) of Various Methods on the BTCV Datasets}
\label{exp_btcv}
\begin{tabular}{@{}lc@{\hspace{15pt}}c@{\hspace{15pt}}c@{\hspace{15pt}}c@{\hspace{15pt}}c@{\hspace{15pt}}c@{\hspace{15pt}}}
\toprule
Methods            & Left Kidney & Right Kidney & Aorta & Spleen &Esophagus &Liver  \\ \midrule
SAM-B (\textit{N} points)\cite{kirillov2023sam}&4.3         & 4.4      & 5.4 & 8.4 & 2.3    & 22.9 \\  
SAM-B (6\textit{N} points)\cite{kirillov2023sam}&  39.9      & 22.6       & 54.5 & 56.3 & 33.2    & 49.2 \\  
SAM-Med3D(6 points)\cite{wang2023sammed3d}      &  \textbf{78.5}      & 61.6    & 37.3 & 55.9 & 36.0     & \textbf{52.4}        \\
\midrule
Ours-MRF (6 points)               & 70.2       & \textbf{68.7}    & \textbf{56.4} & \textbf{58.3} & \textbf{52.9}     & 42.4     \\ \bottomrule
\end{tabular}
\end{table*}

\subsection{Quantitative and Qualitative Results}

Table \ref{tb1} presents the results from various methods after six rounds of interaction on four datasets. In this study, the rationale for presenting the results of the sixth round in the table is to maintain consistency with the Mem3D\cite{zhou2021qualityawaremem} method and to assess the performance level that the model can achieve with fewer interactions. Our proposed method outperforms previous methods in terms of the DSC across four datasets. Compared to Mem3D, our approach shows improvements of 2.4\% and 1.7\% on the MSD-Lung and MSD-Colon datasets, respectively, and enhancements of 0.4\% and 1.0\% on the KiTS-Organ and KiTS-Tumor datasets. These findings highlight the effectiveness of our method in processing various types of medical volumetric images and its capability in achieving accurate segmentation with fewer interactions.

\begin{figure}[!t]
\centering
\includegraphics[width=0.5\textwidth]{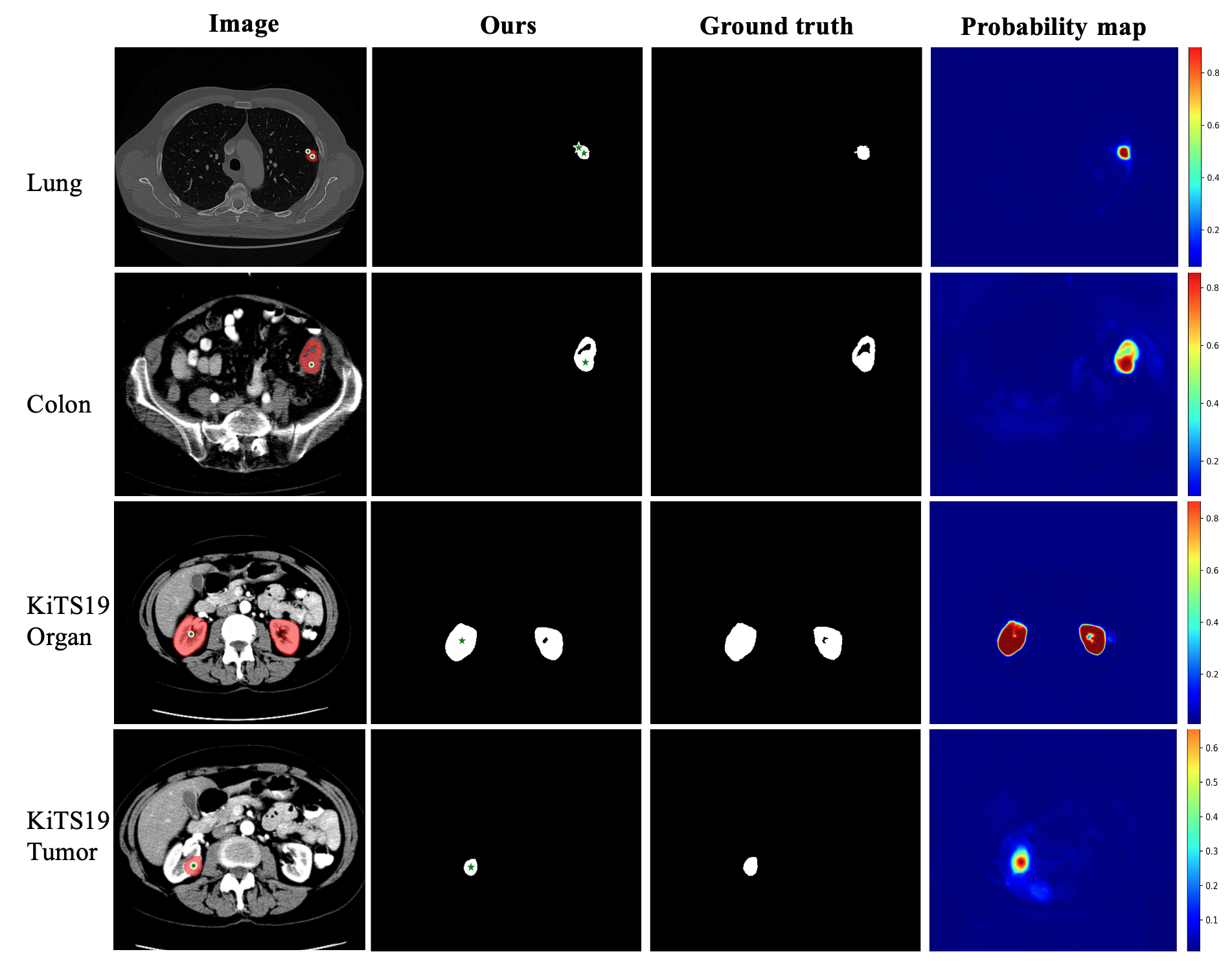}
\caption{Segmentation Results from 2D Interaction Module on Different Datasets. The results serve as prompt frames.}
\label{fig4}
\vspace{-0.5cm} 
\end{figure}

Figure \ref{fig4} illustrates the segmentation results obtained from 2D interactive modules based on user clicks. It can be observed that with only 1 or 2 click interactions, the model can achieve reasonably accurate segmentation of the target objects. The probability map clearly shows that the user click locations have the highest probability values. The model effectively utilizes the user's interactive feedback information to guide the segmentation process and make corrections, demonstrating the advantage of interactive segmentation algorithms.

\textbf{The impact of the number of user interactions on Dice scores:} 
As illustrated in Figure \ref{fig3}, a significant trend is observed in the improvement of the Dice score with increasing interaction rounds. Initially, there is a rapid growth in the Dice score, which gradually decelerates and eventually stabilizes. In the KiTS19-Organ dataset, it has been observed that when the number of interaction rounds exceeds 7, further improvements in segmentation accuracy become relatively limited. In scenarios in which the MRF module was not implemented, the Dice score showed notable fluctuations with the increasing number of interactions. By contrast, using the MRF module led to a consistently increasing trend in the Dice score, thereby demonstrating the efficacy and impact of the MRF module.
\begin{figure*}[!t]
  \centering
  \includegraphics[width=20cm, height=10cm, keepaspectratio]{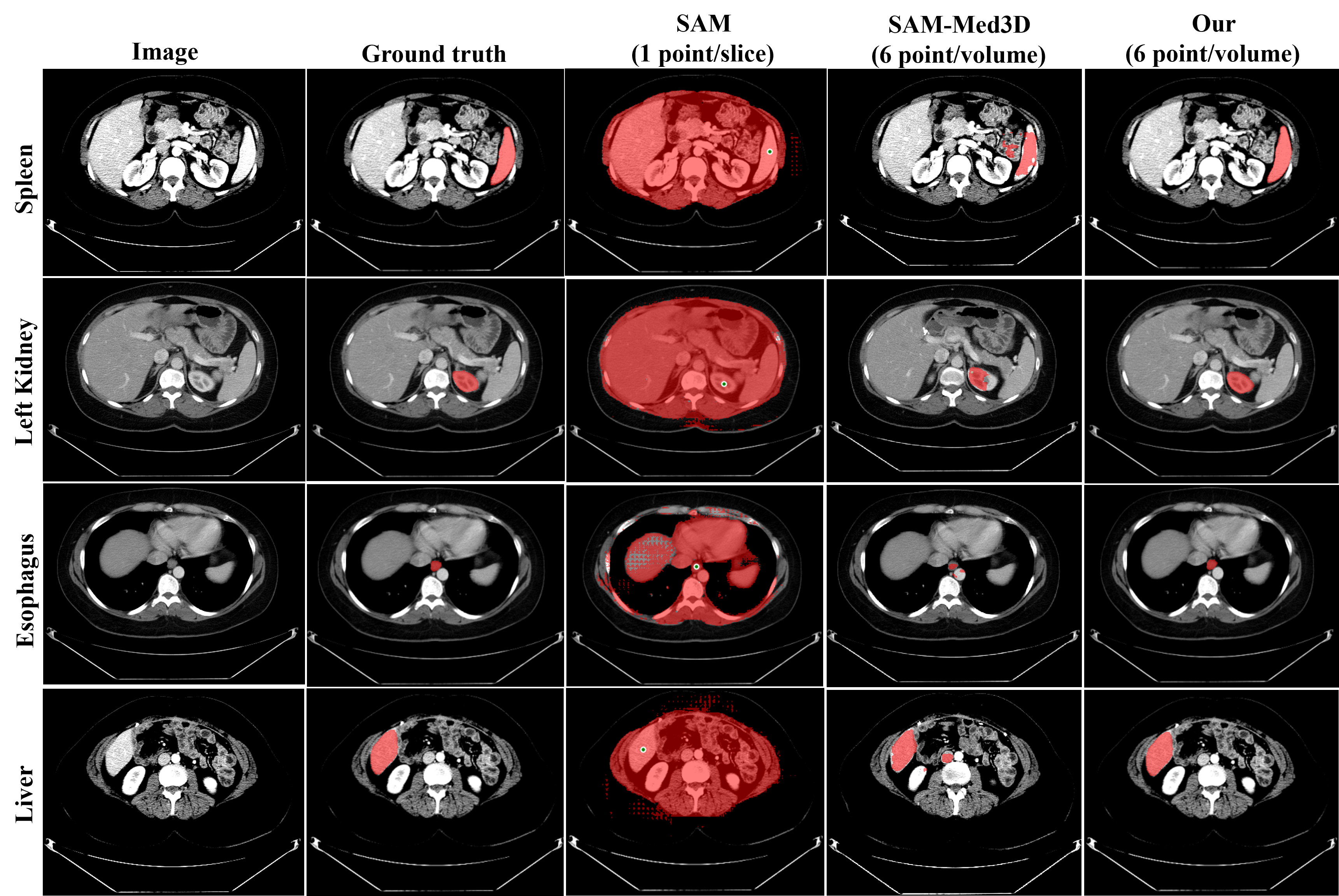}
  \caption{Segmentation results of BTCV dataset}
  \label{btcvpng}
\end{figure*}

\begin{figure}[!t]
\centering
\includegraphics[width=0.48\textwidth]{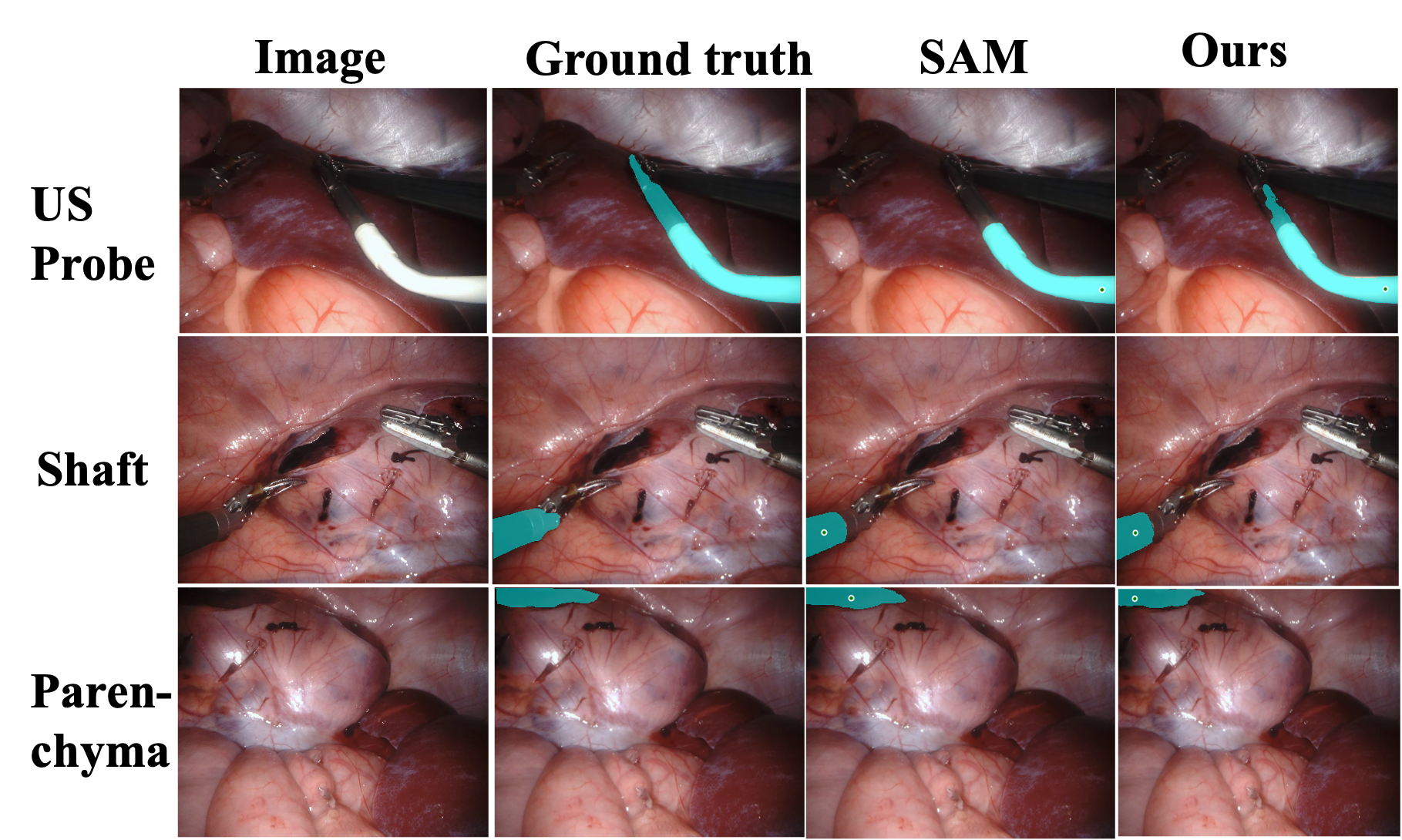}
\caption{Segmentation results of four sequences in EndoVis18 testset}
\label{figendo}
\vspace{-0.5cm} 
\end{figure}
\setlength{\heavyrulewidth}{1.2pt} 
\setlength{\lightrulewidth}{0.6pt} 
 \setlength{\tabcolsep}{10mm}
\begin{table}[ht]
\centering
\caption{Evaluation of Generalization Performance on a Pancreas Tumor segmentation}
\label{tbpancreas}
\begin{tabular}{@{}l@{\hspace{10pt}}c@{\hspace{10pt}}c@{\hspace{10pt}}c@{}}
\toprule
Methods     &Colon Cancer       &Pancreas Tumor\\ \midrule
nnU-Net\cite{isensee2021nnunet}     &43.91        &41.65             \\
TransBTS\cite{wang2021transbts}    &17.05                &31.90             \\
nnFormer\cite{zhou2023nnformer}     &24.28               &36.53           \\
Swin-UNETR\cite{swincvpr2022}  &35.21              &40.57         \\
UNETR++\cite{shaker2022unetr++}      &25.36              & 37.25               \\
3D UX-Net\cite{Lee20223DUA}      &28.50             & 34.83            \\
3DSAM-adapter(automatic)\cite{gong20233dsamadapter} &31.97   & 30.20   \\
\hdashline      
SAM-B(10\textit{N} points)\cite{kirillov2023sam}   &39.14      &30.55                  \\
3DSAM-adapter(10 points)\cite{gong20233dsamadapter} &49.99    &57.47  \\

\midrule
Ours(6 clicks/volume)      &\textbf{60.78}               & \textbf{63.76}      \\ \bottomrule
\end{tabular}
\end{table}

\setlength{\heavyrulewidth}{1.2pt} 
\setlength{\lightrulewidth}{0.6pt} 
\setlength{\tabcolsep}{10mm}
\begin{table*}[ht]
\centering
\caption{Evaluation of Generalization Performance for surgical scene segmentation on EndoVis18 dataset.}
\label{tbendo}
\begin{tabular}{@{}l@{\hspace{9pt}}c@{\hspace{9pt}}c@{\hspace{9pt}}c@{\hspace{9pt}}c@{\hspace{9pt}}c@{\hspace{9pt}}c@{\hspace{9pt}}c@{\hspace{9pt}}c@{\hspace{9pt}}c@{\hspace{9pt}}c@{\hspace{9pt}}c@{\hspace{7pt}}c@{\hspace{7pt}}c@{\hspace{7pt}}c@{\hspace{7pt}}c@{\hspace{7pt}}c@{\hspace{7pt}}@{}}
\toprule
\multirow{2}{*}{Method}  & \multicolumn{4}{c}{Seq 1} & \multicolumn{4}{c}{Seq 2} & \multicolumn{4}{c}{Seq 3} & \multicolumn{4}{c}{Seq 4} \\
\cmidrule(lr){2-5} \cmidrule(lr){6-9} \cmidrule(lr){10-13} \cmidrule(lr){14-17}&Clips  & Needle &  Intestine & \parbox[t]{15pt}{US\\Probe}                      & Intestine & Shaft  & Wrist  &Clasper                       & Intestine & Wrist & Shaft  & Clasper                                                    & \parbox[t]{15pt}{US\\Probe} & Shaft &Parenchyma & Wrist \\
\midrule
NUS\cite{allan2018RoboticScene2020} & 1.5   & 0   & 5.5 & 0      & 21.2 & 85.0 & 38.7 & 41.8     & \textbf{68.8} & 47.6 & 79.8 & 39.8      & 0.0 & 35.2 & 4.7 & 29.7   \\
NCT2\cite{allan2018RoboticScene2020} & 72.9 & 1.9 & \underline{27.0} & 44.2   & \textbf{65.9} & 83.7 & \underline{51.4} & \underline{51.1}     & \underline{67.6} & 62.5 & 78.5 & 76.3      & 22.9 & \textbf{73.0} & 10.6 & \underline{49.9} \\
OTH\cite{allan2018RoboticScene2020} & \textbf{84.9} & 1.4  & 23.1 & 38.9   & 38.1 & \textbf{91.7} & \textbf{52.5} & \textbf{51.2}     & 15.3 & \textbf{73.3} & \textbf{92.2} & \textbf{81.3}      & 17.5 & \underline{69.3} & \textbf{26.4} & \textbf{51.3} \\
UNC\cite{allan2018RoboticScene2020} & 77.1 & 0.0  & 22.4 & 42.8   & \underline{59.0} & \underline{91.6} & 46.0 & 46.0     & 76.3 & \underline{68.0} & \underline{91.2} & \underline{79.4}      & 18.3 & \underline{69.3} & 8.4 & 46.9  \\    
SAM-B(\textit{N} points)\cite{kirillov2023sam} & 33.6 & \underline{3.2}  & \textbf{30.6} & \underline{53.8}    & 54.5 & 59.2 & 46.9 & 17.5     & 8.9 & 31.6 & 56.7 & 17.0      & \underline{56.6} & 40.3 &12.7  & 16.9 \\ 
\midrule

Ours(6 points) & \underline{78.6} & \textbf{9.0} & 23.8 & \textbf{61.8}        & 54.5 & 65.4 & 32.3 & 12.0      & 65.6 & 45.9 & 43.5 & 19.5          & \textbf{71.9} &15.9 &\underline{13.5} &11.2 \\ 
\bottomrule
\vspace{0.01em}
\end{tabular}
{\footnotesize 
\parbox{\linewidth}
{
\textit{*Note: The \textbf{values in bold} denote the best results, while \underline{underlined values}  indicate the second best. Table \ref{tbendo}  shows the segmentation performance comparison on the EndoVis18 surgical video dataset. The evaluation is conducted across four surgical sequences (Seq 1-4), each containing different combinations of surgical instruments (Clips, Needle, Shaft, Wrist, Clasper, US Probe) and anatomical structures (Intestine, Parenchyma). The performance metrics are presented as Dice. Five baseline methods (NUS, NCT2, OTH, UNC, and SAM-Base) are compared against our approach.}
}
}
\end{table*}

\textbf{Runtime and GPU Memory Utilization Analysis:} Inference Speed and GPU Memory tests on a single Tesla V100-SXM2 (16GB VRAM), using 100 medical images (480x480 resolution). The model achieved 31.25 FPS and utilized only 2.7 GB GPU memory, indicating superior performance for real-time image segmentation. These results highlight the model's advantages in number of parameters and computational efficiency.

\textbf{Generalization ability assessment:} 
Table\ref{exp_btcv} 
We evaluated our method's generalization ability using the unseen BTCV dataset. The experimental results are shown in Table \ref{exp_btcv} and Figure \ref{btcvpng}. It can be observed that although our model was not trained on the BTCV dataset, it still achieved strong performance on most segmentation tasks. Compared with SAM-B\cite{kirillov2023sam} and SAM-Med3D\cite{wang2023sammed3d}, our method has significant improvements in the DSC metrics of the right kidney, esophagus, aorta, and spleen, demonstrating commendable generalization ability.
Particularly in the esophagus segmentation task, our method surpasses other methods with a DSC value of 52.9\%, indicating its proficiency in capturing the shape and boundary information on unseen data. It is worth noting that the SAM-B required numerous input points to achieve satisfactory performance. 
With \textit{N} points, SAM-B's DSC values for all organs are significantly lower than other methods, likely due to ambiguity introduced by insufficient input\cite{cheng2023sammed2d}. This can lead to incorrect segmentation, such as segmenting the entire image instead of specific anatomical structures. 
As shown in the Figure \ref{btcvpng}, in the absence of sufficient constraints, the model tends to generate unreasonable segmentation results. Increasing the input to 6\textit{N} points substantially improves SAM-B’s performance, effectively reducing ambiguity and guiding the model to generate more accurate results.
However, SAM-B's performance remained inferior to our method except for the liver and Left kidney. In contrast, our method and SAM-Med3D only require 6 prompt points to achieve better segmentation effects than SAM-B. The results on the BTCV dataset prove the generalization ability of our proposed method. Our method can effectively capture the shape and boundary information of organs with a small number of prompt points, achieving better effects than SAM-B. The results highlight its robust adaptability to unseen medical image data, offering a promising solution for medical image annotation tasks.
Additionally, to further validate the performance of our approach, we tested the model with four methods from the 2018 Robotic Scene Segmentation Challenge\cite{allan2018RoboticScene2020}. These four methods were proposed by research teams from the National University of Singapore (NUS), the National Center for Tumor Diseases (NCT), the Ostbayerische Technische Hochschule Regensburg (OTH), and the UNC, respectively. By comparing our method with these top-performing approaches in the challenge, we can objectively assess the generalization ability of our algorithm. As illustrated in Table \ref{tbendo}, our proposed method demonstrates remarkable segmentation performance in the surgical video segmentation task, specifically without relying on EndoVis18 surgical video datasets for training.

In several categories, such as Clips, US Probe, Intestine and Needle, our approach achieves results comparable to, or even surpassing, other methods specifically designed and trained for surgical scene segmentation using surgical video data. This  performance highlights the strong generalization capability of our methodology. Notably, in the Seq1-Needle and Seq4-USProbe categories, our method achieves DSC of 9 and 71.9, respectively, compared to the NCT2 method's scores of 1.9 and 22.9.
It proves that our method enables effective transfer and application of features learned from other image domains to the new surgical video domain. Despite not fully reaching the existing best level in categories like Seq1-Clips, Seq2-Intestine, Seq3-Intestine, and Seq4-Parenchyma, the performance gap is relatively minor. Nevertheless, our method exhibits relatively inferior performance in categories such as Seq3-Clasper, Seq2-Clasper, and Seq4-Wrist. This could be attributed to the unique visual characteristics of these targets in surgical videos, potential variations among different surgical video sequences, and the domain gap between our previous training data and the surgical video domain. Considering that the existing comparative methods are explicitly designed for surgical scene segmentation, while our approach achieves comparable performance without any fine-tuning on surgical data, these results are acceptable.

Overall, although our method is not trained on the EndoVis18 dataset, its performance in multiple categories is already close to or even surpasses that of the compared surgical scene segmentation methods, demonstrating the excellent generalization ability and universality of our proposed framework.

As shown in Table \ref{tbpancreas}, \textbf{in the pancreas tumor segmentation task} using the MSD-Pancreas dataset, our model achieved a Dice score of 63.76\%, demonstrating its ability to accurately segment tumors.
This performance is particularly considering the difficulty in segmenting pancreas tumors due to their ambiguous borders.
Compared to the 3DSAM-adapter methods, our approach exhibits improvements in pancreas tumor segmentation, with a 10.9\% increase in Dice score. By training the model on a diverse range of data distributions, it acquired a high degree of generalization, resulting in competitive performance even on the challenging and previously unseen pancreas tumor dataset.

\subsection{Ablation Study}
To evaluate the effectiveness of the proposed MRF Module in our interactive segmentation framework, we conducted an ablation study. 
We performed experiments on four CT volume datasets, comparing the performance of two approaches: one that directly uses the current round segmentation results as output (denoted as w/o MRF Module) and another that utilizes our proposed MRF module to obtain the optimal results (denoted as w/ MRF Module).

After each round of interaction, we computed the average Dice score on the segmented slices. Table \ref{tb5} presents the Dice scores for both scenarios. The results indicate that solely relying on the current round of segmentation outcomes can lead to inconsistent performance. 
This inconsistency can be attributed to the potential degradation of segmentation quality in some slices when all slices are updated based solely on single-slice refinement. In contrast, the segmentation results obtained using the MRF module outperformed the baseline by 0.52\%, 5.19\%, 0.36\%, and 5.58\% in terms of Dice score on the four datasets, respectively. These improvements demonstrate the effectiveness of the MRF module in selecting higher-quality masks from the mask sequences generated in two rounds of segmentation and combining them into a new sequence. 
This strategy avoids the negative impacts caused by lower-quality masks in the current sequence. 

In summary, the ablation study highlights the importance of the MRF module in our interactive segmentation framework. By selectively choosing the superior segmentation results for each slice from multiple rounds, the MRF module enhances the consistency, reliability, and accuracy of the segmentation output, ultimately reducing the manual annotation effort required from users when working with medical images.

\section{CONCLUSION}
This paper proposes a novel 3D interactive medical image segmentation framework that incorporates a 2D interaction network, MRF module, and a Memory model. The MRF Module enhances segmentation outcomes by merging high quality masks from successive rounds, thereby improving accuracy and reducing errors. Our framework does not necessitate high-performance computing resources and can be implemented on devices with limited computational capabilities. This method reduces user interactions, decreasing the time needed for annotators and accelerating the annotation process for 3D medical image datasets.
\section*{References}
\bibliographystyle{IEEEtran}
\bibliography{main}
\end{document}